\documentclass[11pt]{article}

\usepackage[]{acl}

\usepackage{times}
\usepackage{latexsym}

\usepackage[T1]{fontenc}
\usepackage{booktabs}
\usepackage{multirow}
\usepackage{amsmath}
\usepackage{bbding}
\usepackage{tikz-dependency}
\usepackage[normalem]{ulem}
\usepackage{amsfonts}
\usepackage[utf8]{inputenc}
\usepackage{adjustbox}
\usepackage{subcaption}
\usepackage{microtype}

\usepackage{multirow}
\usepackage[HTML]{xcolor}
\usepackage{enumitem}

\newcommand{\StdFunc}{\textbf{\textsc{\textcolor[HTML]{4D4D4D}{Standard Function}}}}

\newcommand{\NaturalFunc}{\textbf{\textsc{\textcolor[HTML]{4D4D4D}{NaturalFunction}}}}

\newcommand{\PhraseDep}{\textbf{\textsc{\textcolor[HTML]{4D4D4D}{PhraseDependency}}}}
\newcommand{\AtBound}{\textbf{\textsc{\textcolor[HTML]{4D4D4D}{AtBoundary}}}}
\newcommand{\NoFunc}{\textsc{\textcolor[HTML]{CF597E}{NoFunction}}}
\newcommand{\MoreFunc}{\textsc{\textcolor[HTML]{E47D6C}{MoreFunction}}}
\newcommand{\FiveFunc}{\textsc{\textcolor[HTML]{EACC83}{FiveFunction}}}

\newcommand{\RandDep}{\textsc{\textcolor[HTML]{B4D388}{RandomDep}}}
\newcommand{\BigramDep}{\textsc{\textcolor[HTML]{43B58A}{BigramDep}}}
\newcommand{\CrossBound}{\textsc{\textcolor[HTML]{0082B9}{WithinBoundary}}}

\newcommand{\fw}[1]{\texttt{\colorbox[HTML]{FFF2B2}{#1}}}

\newcommand{\posdelta}[1]{\textcolor{blue!70}{\tiny(#1)}}
\newcommand{\negdelta}[1]{\textcolor{red!70}{\tiny(#1)}}

\usepackage{inconsolata}

\usepackage{graphicx}

\title{Function Words as Statistical Cues for Language Learning}

\author{
Xiulin Yang \quad Heidi Getz \quad Ethan Gotlieb Wilcox\\
Georgetown University\\
\texttt{\{xy236, heidi.getz, ethan.wilcox\}@georgetown.edu}
}

\begin{document}
\maketitle
\begin{abstract}
What statistical properties might support learning abstract grammatical knowledge from linear input? We address this question by examining the statistical distribution of function words. Function words have been argued to aid acquisition through three distributional properties: high frequency, reliable syntactic association, and phrase-boundary alignment. We conduct a cross-linguistic corpus analysis of 186 languages, which confirms that all three properties are universal. Using counterfactual language modeling and ablation experiments on English, we show that preserving these properties facilitates acquisition in neural learners, with a Goldilocks effect: function words must be frequent enough to be reliable, yet diverse enough to remain informative to structural dependency. Probing analyses further reveal that different learning conditions produce systematically different reliance on function words.\footnote{Code: \url{https://github.com/picol-georgetown/function_word}; models: \url{https://huggingface.co/collections/xiulinyang/function-words}.}
\end{abstract}

\section{Introduction}
A central puzzle in language acquisition is how learners abstract grammatical knowledge from linear input. Extensive research on statistical learning argues that human language learners make use of distributional information in their input, such as transition probability \citep{thompson2007statistical} and prosodic grouping \citep{morgan1987structural,romberg2010statistical}.
Within this literature, function words, such as determiners, auxiliaries, and prepositions, have been argued to play a critical role \citep[e.g.,][]{green1979necessity, hicks2006impact}. Three properties of function words have been argued to facilitate acquisition: 
(i) high lexical frequency, (ii) reliable association with particular syntactic structures, and (iii) consistent position at phrase boundaries. Because of these properties, function words might serve as anchor points for learning, cuing particular word sequences for analysis (e.g., the Anchoring Hypothesis; \citealt{valian1988anchor}; see also \citealt{morgan1987structural} on boundary-aligned cues) or giving learners a small set of high-frequency items to track (e.g., the Marker Hypothesis; \citealt{green1979necessity}, with related evidence from \citealt{getz2019privileged,zhang2015grammatical,mintz2006finding}). 

This literature suggests that function words contribute in important ways to statistical learning, yet two major gaps remain. 
First, existing studies draw conclusions about the properties of function words primarily from English or a small set of languages \citep[e.g.,][]{green1979necessity,kimballSevenPrinciplesSurface1973,ClarkClark1977,shi1998phonological}, limiting the generalization of their findings. Linguists have characterized the grammatical properties of function words in detail for thousands of languages \citep{haspelmath2005world}, and these analyses suggest that many of the distributional properties of function words are likely to be universal; however, this has not been confirmed through statistical analyses of diverse language corpora. If these properties play a systematic role in language acquisition in general, it is important to examine whether they are robust across languages (RQ1). Second, because these properties are often examined in isolation in simplified artificial language settings, it remains unclear whether their findings can be extended to more complex and therefore ecologically valid learning simulations (RQ2.1) and how and to what extent each property contributes to syntactic generalization (RQ2.2).
Furthermore, is the role of function words as statistical cues in language learning carried over to language processing? Or are other cues used to parse sentences once grammatical representations become robust enough? (RQ3)

We answer RQ1 by conducting a cross-linguistic analysis using data from the Universal Dependencies (UD) project \citep{de-marneffe-etal-2021-universal}, confirming the three distributional properties are universal across 186 languages in our sample. To address RQ2, we take transformer language models (LMs) as domain-general and weakly-biased learners \citep{wilcox2024using} and train them on counterfactual variants of natural text in which each distributional property of function words is systematically manipulated. We also train $n$-gram models as baselines for syntactic generalization. We confirm that a language is most learnable when all three statistical properties of function words jointly hold. Additionally, we observe a Goldilocks effect, whereby function words must be frequent enough to be reliable, yet sufficiently diverse to remain informative for their linguistic dependents. Finally, for RQ3, we conduct attention probing and function word ablation experiments to examine how LMs internally represent and deploy function word information. We find models trained on languages with the three function word properties develop attention heads specialized for function words and rely more on function word information during processing.

\section{Background \& Related Work}

\subsection{Defining Function Words}
The distinction between function and content words is foundational in linguistic theory \citep{rizzi2016functional,abney1987english} and language development research \citep{dyeLexicalFunctionalCategories2019}, though the boundary is not always clear-cut \citep{kayne2005movement}\footnote{``The conclusion is that
the number of functional elements in syntax is not easy to estimate, but at the same
time that 100 would be a low estimate.''\citep[][p.288]{kayne2005movement}}. Typically, function words are characterized as a closed class with high frequency \citep{morgan1987structural}, reduced prosodic prominence \citep{selkirk2014prosodic,languages6040197}, systematic alignment with phrase boundaries \citep{kimballSevenPrinciplesSurface1973,christophe2008bootstrapping,ClarkClark1977}, and relatively light semantic content despite high grammatical utility \citep{carlson1983marking}.
In our modeling experiments, function words are defined purely through their phrasal alignment and high frequency. 
We leave cues such as semantic content and prosodic prominence as avenues for future research.

\subsection{Function Words in Language Acquisition}
Although early speech production by children usually lacks function words \citep{brown2013first}, evidence suggests that even in infancy, learners perceive these elements \citep{shi2006recognition,shi1999newborn} and are sensitive to their distribution  \citep{hochmann2010word,gerken2005infants,christophe2008bootstrapping, kedar2006getting}. Experiments with artificial languages have shown that languages are more learnable by humans when function words are (1) more frequent than content words \citep{valian1988anchor}; (2) reliable predictors of specific phrase structures \citep{green1979necessity, getz2019privileged}; and (3) systematically positioned at phrase boundaries \citep{morgan1987structural}.


\subsection{Computational Perspectives}

Our work aligns with a growing body of research examining how manipulating training data shapes language model behavior. One line of work alters the overall statistical distribution of languages, testing whether languages that are less aligned with hypothesized grammatical principles are harder for models to learn \citep{kallini-etal-2024-mission,yang-etal-2025-anything,ziv-etal-2026-biasless}. Another line manipulates the proportion of certain targeted constructions, often referred to as Filtered Corpus Training \citep{patil-etal-2024-filtered}, and has been applied to a range of phenomena, including AANN constructions \citep{misra-mahowald-2024-language}, \textit{let alone} \citep{scivetti-etal-2025-unpacking}, and Poverty of the Stimulus effects \citep{yang2026unified}, among others \citep[e.g.,][]{yao2025both}.

In contrast, computational work on function words remains relatively limited and falls into two strands. The first uses computational models to simulate human acquisition, showing that distributional cues such as frequent frames can support syntactic categorization \citep{mintz2003frequent, chemla2009categorizing, gutman2015bootstrapping}. Related work demonstrates that placing function words at phrase boundaries facilitates categorization and word learning \citep{Mintz2002-MINTDS,johnson-etal-2014-modelling}. More recently, \citet{ma2025implicit} replicate \citet{valian1988anchor} with LLMs and find limited sensitivity to marker frequency in artificial language settings. The second strand probes pretrained models, with mixed evidence on whether language or vision-LMs encode knowledge of function words \citep{kim-etal-2019-probing, ettinger-2020-bert, portelance2024learning}.

\begin{table*}[t]
\footnotesize
\centering
\begin{tabular}{lllll}
\toprule
\textbf{Properties} & \textbf{Manipulation} & \textbf{Example} \\
\midrule

\multirow{4}{*}{Lexical Frequency} 
& \NoFunc\ &  dog happily chasing dog garden .\\

& \MoreFunc\ &  \fw{thi} dog \fw{wist} happily chasing \fw{que} dog \fw{ap} \fw{tho} garden .\\

& \FiveFunc\ &\fw{the} dog \fw{will} happily chasing \fw{the} dog \fw{at} \fw{the} garden . \\

& \StdFunc\ &  \fw{a} dog \fw{is} happily chasing \fw{another} dog \fw{in} \fw{the} garden .  \\
\midrule

\multirow{3}{*}{Structural Dependency} 
& \RandDep &  \fw{by} dog \fw{in} happily chasing \fw{at} dog  \fw{by} \fw{will} garden .  \\

& \BigramDep &  \fw{by} dog \fw{in} happily chasing \fw{by} dog  \fw{at} \fw{will} garden . \\

& \PhraseDep & \fw{a} dog \fw{is} happily chasing \fw{another} dog \fw{in} \fw{the} garden .  \\
\midrule

\multirow{2}{*}{Phrase boundary} 
& \CrossBound  & \fw{a} dog happily \fw{is} chasing \fw{another} dog  \fw{the} \fw{in} garden .  \\

& \AtBound &  \fw{a} dog \fw{is} happily chasing \fw{another} dog \fw{in} \fw{the} garden .  \\

\bottomrule
\end{tabular}
\caption{An overview of manipulated languages}
\label{Tab:languages}
\end{table*}

\section{Do Function Words Share Universal Distributional Properties?}
Research on function words in the field of language acquisition has been heavily skewed to English. In this section, we take a broader crosslinguistic perspective. We ask whether the properties identified as crucial for learning, i.e., high-frequency, structural predictability, and phrase boundary alignment, are linguistic universals. For this purpose, we use UD (v2.17), which includes 339 treebanks across 186 languages.. 

\begin{figure}[t]
    \centering
    \begin{subfigure}[t]{0.49\linewidth}
        \centering
        \includegraphics[width=\linewidth]{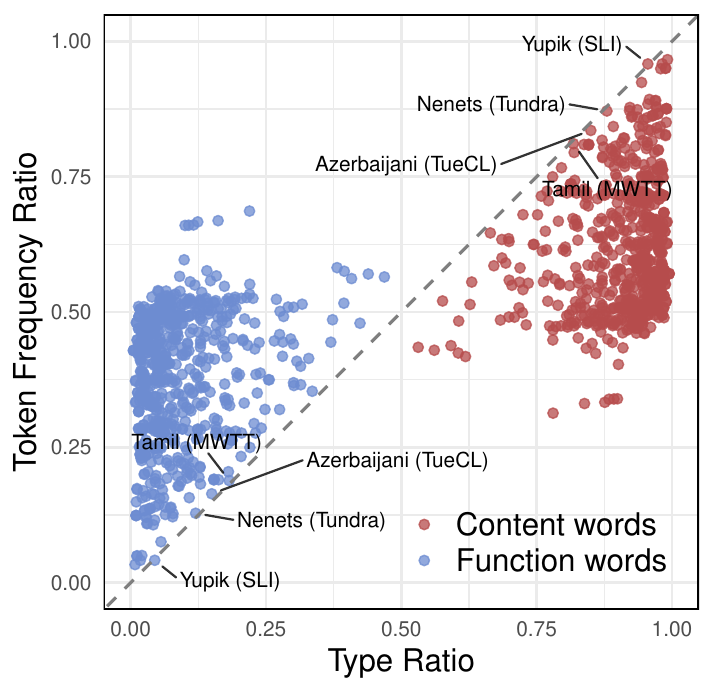}
        \caption{Word frequency ratio vs.\ type ratio}
        \label{fig:freq}
    \end{subfigure}
    \hfill
    \begin{subfigure}[t]{0.49\linewidth}
        \centering
        \includegraphics[width=\linewidth]{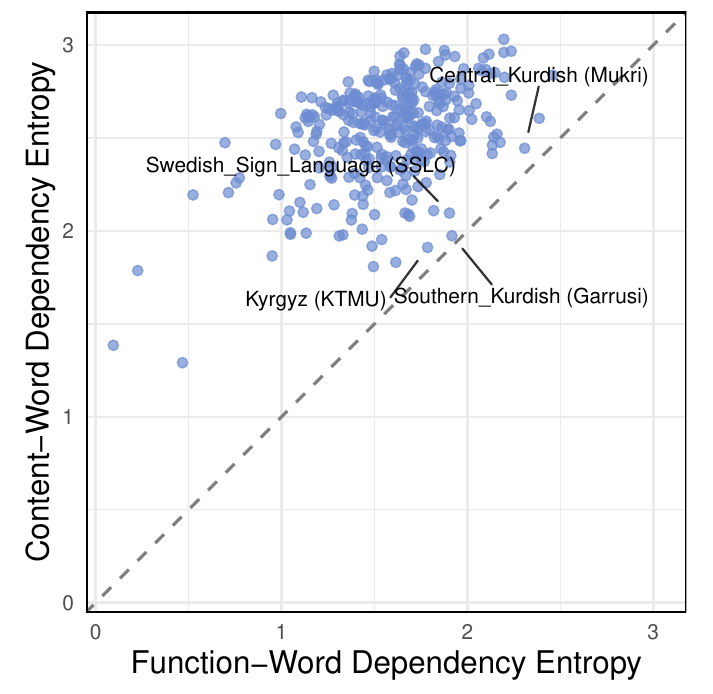}
        \caption{Function vs.\ Content word dependency entropy}
        \label{fig:entropy}
    \end{subfigure}
    \caption{Distributional properties of function words across languages. Function words tend to have higher frequency-to-type ratios and lower syntactic dependency entropy than content words. The top four languages (with treebank names in parentheses) closest to the diagonal are labeled.}
    \label{fig:functionword-ud}
\end{figure}

\paragraph{High frequency}
\label{freq}
We quantify the frequency of function words by jointly considering their inventory size (\textit{types}) and their usage frequency (\textit{tokens}). Specifically, for a word class $c$, we define the \emph{type ratio} as $\tfrac{|\mathcal{V}_c|}{|\mathcal{V}|}$, where $\mathcal{V}_c$ denotes the set of unique word types belonging to class $c$ (content or function words) and $\mathcal{V}$ the full vocabulary. 
We further define the \emph{token frequency ratio} as $\tfrac{\sum_{w \in \mathcal{V}_c} \text{count}(w)}{\sum_{w \in \mathcal{V}} \text{count}(w)}$.  Following the UD convention, we define function words as closed-class items excluding \textsc{num} and treat \textsc{num} and other open-class categories as content words. We exclude non-linguistic tags (\textsc{x}, \textsc{punct}, \textsc{sym}).

Results are shown in Figure~\ref{fig:freq}, where each UD treebank contributes two points, corresponding to scores of function words and content words. If word classes were uniformly distributed, the proportion of function word tokens would match their type proportion, yielding points along the diagonal. Instead, we observe a robust cross-linguistic pattern: function words occupy a small inventory size but account for a disproportionately large share of tokens, while content words fall below the diagonal across languages, reflecting larger inventories with lower token frequencies.

\paragraph{Reliable structural prediction} 
Function words occur in predictable syntactic environments. For instance, in English, determiners occur only at the beginning of noun phrases. To test this cross-linguistically, we analyze the \textit{syntactic selectivity} of POS tags by modeling dependency trees as undirected graphs. We calculate the dependency entropy of the syntactic neighbors (i.e., co-dependents) for each POS category. 
Let $X$ and $Y$ be nodes that are in a syntactic dependency with each other, and ranging over a syntactic tagset $\mathcal{T}$.
We calculate the syntactic selectivity as the conditional entropy:
\begin{equation}
    H(X \mid Y=y) = -\sum_{x \in \mathcal{T}} p(x|y) \log_2 p(x|y)
\end{equation}

where $p(x|y)$ is the probability that a connected node has the POS tag $x$, given the base node has tag $y$. Probabilities are estimated via maximum likelihood estimation from our UD corpus. We then calculate the average entropy for a set of function word tags $\mathcal{F} \subset \mathcal{T}$,
\begin{equation}
    \bar{H}_{\mathcal{F}} = \frac{\sum_{f\in \mathcal{F}} \text{Freq}(f) H(X\mid Y = f)}{\sum_{f\in \mathcal{F}} \text{Freq}(f)}
\end{equation}

\noindent and likewise calculate this metric for content words $\bar{H}_\mathcal{C}$. Figure~\ref{fig:entropy} illustrates the results. We observe a consistent pattern across languages: $\bar{H}_{\mathcal{F}}$ is always lower than $\bar{H}_{\mathcal{C}}$. This indicates that function words exhibit high syntactic selectivity, connecting to a restricted set of categories, whereas content words operate as high-entropy hubs with diverse connections.

\paragraph{Phrase Boundary Alignment} 
\begin{figure}[t]
\centering
\begin{adjustbox}{max width=\linewidth}
\begin{dependency}
\begin{deptext}[column sep=0.5em]
\textcolor{teal}{[}a \& dog\textcolor{teal}{]} \& 
\textcolor{orange}{[}is \& happily \& chasing \& 
\textcolor{teal}{[}another \& dog\textcolor{teal}{]} \& 
\textcolor{purple}{[}in \& 
\textcolor{teal}{[}the \& garden\textcolor{teal}{]}
\textcolor{purple}{]}
\textcolor{orange}{]} \\
\end{deptext}

\deproot{5}{root}
\depedge{2}{1}{det}
\depedge{5}{2}{nsubj}
\depedge{5}{3}{aux}
\depedge{5}{4}{advmod}
\depedge{7}{6}{det}
\depedge{5}{7}{obj}
\depedge{10}{8}{case}
\depedge{10}{9}{det}
\depedge{5}{10}{obl}
\end{dependency}
\end{adjustbox}
\caption{An example dependency tree with constituent spans highlighted using colored brackets: \textcolor{teal}{NP}, \textcolor{orange}{VP}, and \textcolor{purple}{PP}. Phrase labels are theoretically neutral.}
\label{fig:dependency-example}
\end{figure}

\begin{figure}[t]
    \centering
    \includegraphics[width=\linewidth]{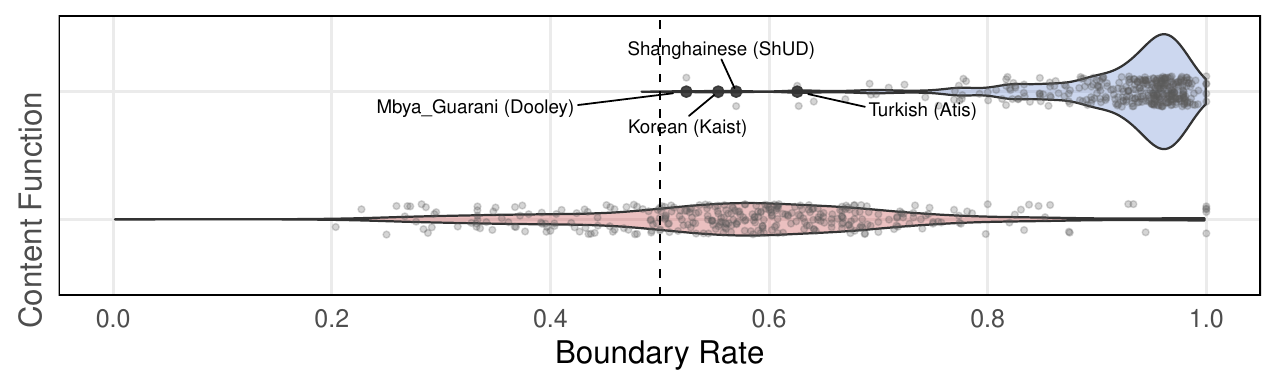}
    \caption{The boundary ratio of function and content words across languages. Function words consistently show higher boundary ratios across languages.}
    \label{fig:boundary}
\end{figure}

We examine the degree to which words belonging to class $c$ align with phrase boundaries ($\frac{\sum_{w \in \mathcal{V}_c} \mathbf{1}\{\text{$w$ is at a boundary}\}}{|\mathcal{V}_c|}$). While a constituency-parsed dataset would be preferable, using one would require drastically reducing the number of languages considered. We therefore opt to use UD as it provides the most comprehensive collection of multilingual syntactic data currently available. We approximate constituent spans using the yield of dependency subtrees: For each target word, we identify the subtree governed by its syntactic head and check if the word appears at the subtree's left or right periphery.

We apply two refinements to this heuristic. First, dependency trees do not perfectly map to constituent structures. For example, as shown in Figure~\ref{fig:dependency-example}, the auxiliary \textit{is} depends on the verb \textit{chasing}. A naive subtree yield of the verb phrase would place \textit{is} internally (as \textit{dog} is also a dependent to \textit{chasing}), failing to capture its boundary status in the corresponding constituent. To mitigate such alignment errors, we exclude categories that often function as dependents of predicates (specifically \textsc{pron}, \textsc{part}, and \textsc{aux}) and focus our analysis on \textsc{adp}, \textsc{det}, \textsc{sconj}, and \textsc{cconj}. For content words, we only focus on \textsc{adj} and \textsc{num} as other POSes are all related to verbs. 

Second, we adjust for nested structures, such as an \textsc{np} embedded within a \textsc{pp} (e.g., \textit{in [the garden]}). In standard UD, the adposition \textit{in} and the determiner \textit{the} both attach to the noun \textit{garden}. A naive boundary check on the subtree of \textit{garden} would identify \textit{in} as the boundary, incorrectly labeling \textit{the} as internal. To address this, we implement a relaxation rule: if a word is immediately preceded (or followed) by another word from the same function/content word class that marks the constituent boundary, we also consider the target word as a boundary marker.\footnote{Although these adjustments reflect English-centric assumptions, they serve as a necessary approximation to maintain methodological consistency across diverse languages lacking gold-standard constituency resources.}

Figure~\ref{fig:boundary} shows a strong cross-linguistic tendency: the median ratio of function words at phrase boundaries is 0.95 (minimum 0.55 in Korean), compared to only 0.58 for content words.

\section{Which Properties of Function Words Support Syntactic Learning?}

Having established that the three distributional properties of function words are cross-linguistically robust, we ask whether findings from human artificial language learning generalize to natural language and which of these properties matters most. 

\subsection{Training Data Construction}
We construct our training corpora by manipulating text from Wikipedia.\footnote{\url{https://huggingface.co/datasets/wikimedia/wikipedia}} We use Wikipedia instead of BabyLM \citep{conll-2024-babylm} due to the lack of accurate parsers for child-directed speech (CDS), as one of our manipulations (\CrossBound)~ requires reliable identification of function words and their syntactic heads.\footnote{One potential concern is that Wikipedia may exhibit a different distribution of function words compared to CDS, which could limit the generalizability of our findings to language acquisition. However, we find that Wikipedia and UD-CHILDES \citep{yang-etal-2025-ud} have comparable proportions of function words under the \CrossBound~ transformation (CHILDES: 54\%; Wikipedia: 55\%).} All text is lowercased prior to training. We use Stanza \citep{qi2020stanza} to obtain word-level POS tags and dependency parses for identifying function words and phrase boundaries during corpus manipulation.
\subsection{Function Word Identification}
Same as Experiment 1, we identify function words based on closed-class POS tags in UD. Specifically, we include \textsc{det} (determiners), \textsc{adp} (adpositions), \textsc{cconj} (coordinating conjunctions), \textsc{sconj} (subordinating conjunctions), and \textsc{aux} (auxiliaries).

We explicitly exclude pronouns, degree quantifiers, and numerals. Pronouns often realize core argument positions; manipulating them would directly alter the argument structure of sentences, introducing a confound where degraded performance reflects missing arguments rather than missing structural cues. Numerals and degree quantifiers are excluded as they exhibit more variable and semi-lexical behavior across languages: they are often gradable, can function as independent predicates, and in many languages pattern more like adjectival or nominal modifiers rather than core determiners \citep{corver2013semi}. 

To construct the inventory, we collect word types associated with the selected POS tags from the GUM \citep{zeldes2017gum} and EWT \citep{silveira14gold} treebanks. After filtering items with fewer than 10 occurrences and obvious annotation errors, we obtain a final set of 116 English function words (see Appendix~\ref{funcw}).

\subsection{Experimental Conditions}
Building on prior work on distributional properties of function words \citep{valian1988anchor,green1979necessity,morgan1987structural,christophe2008bootstrapping}, we generate manipulated corpus variants targeting function word properties that have been hypothesized to contribute to their status as cues for grammatical structure. To ensure comparability across conditions, content words and total dataset size are held constant; sentence length distributions are also matched across conditions (except \NoFunc). All experiments use English only (see Section~\ref{limitation} for discussion).

\paragraph{High Lexical Frequency.}
We manipulate lexical frequency by varying how token counts are distributed across function word types, while holding the total number of function word tokens constant (except in \NoFunc). Specifically, we vary the size of the function word inventory to induce variation in their lexical frequency:
\begin{itemize}[leftmargin=1em]
\setlength\itemsep{-0.3em}
    \item \StdFunc: The natural English inventory (116 types).
    \item \NoFunc: A language with all function words removed.
    \item \FiveFunc: A minimized inventory where all function words within a syntactic category (e.g., all determiners) are mapped to a single type (5 function word types in total).
    \item \MoreFunc: An expanded inventory where each natural function word is mapped to 10 distinct pseudowords based on their forms using Wuggy \cite{keuleersWuggyMultilingualPseudoword2010a}, increasing the inventory size to $\sim$1.2k types.
\end{itemize}

\paragraph{Structural Predictability.} We manipulate the statistical dependency between function words and their context:
\begin{itemize}[leftmargin=1em]
\setlength\itemsep{-0.3em}
    \item \PhraseDep: The natural baseline, where function words reliably predict specific phrase structures.
    \item \BigramDep: Function words are determined by the \textit{following} word (i.e., the local bigram dependency) rather than by structural class.\footnote{We first construct a one-to-one mapping between vocabulary items plus a special unknown word and function words, keeping the number of function word types the same. When regenerating the corpus, function words are generated conditioned on the following word.}
    \item \RandDep: Function word location is preserved, but their identity is randomly shuffled.
\end{itemize}

\paragraph{Phrase-boundary Alignment.} We manipulate the structural placement of function words:
\begin{itemize}[leftmargin=1em]
\setlength\itemsep{-0.3em}
    \item \AtBound: The natural baseline, where function words appear systematically at phrase boundaries.
    \item \CrossBound: Function words are displaced from phrase boundaries to positions immediately adjacent to their syntactic heads. This minimizes dependency length but destroys boundary cues. This manipulation changes 55\% of the location of function words in over 99\% of the sentences in the training split. 
\end{itemize}

As each baseline corresponds to natural English, we refer to this shared unmanipulated condition as \NaturalFunc. Examples are shown in Table~\ref{Tab:languages}, and the effects of each manipulation on the three properties are summarized in Appendix~\ref{condition}. 

\subsection{Model \& Training}
We train GPT-2 Small models from scratch on each language variant. To ensure a fair comparison, a dedicated tokenizer is trained for each manipulated corpus to accommodate vocabulary changes. Each model is trained for 10 epochs. We report results averaged over 3 random seeds. Detailed hyperparameters are provided in Appendix~\ref{trainingdetails}. For each language, we also train a $5$-gram model with Kneser-Ney smoothing using KenLM \citep{heafield2011kenlm}.

\subsection{Evaluation}
We do not use perplexity as an evaluation metric because our manipulations alter the entropy of the corpora. Instead, we evaluate syntactic generalization using the BLiMP benchmark \citep{warstadt-etal-2020-blimp-benchmark}.\footnote{To make sure that Transformer models are able to learn the manipulated languages, we measured language model competence by asking each model to perform a 7-way forced-choice task on sentences consistent with its own training distribution vs. distractors, which tests whether it has learned the underlying grammar of its variant. The result is reported in Appendix~\ref{language_competence}.} Manipulated examples are shown in Table~\ref{minimalpair}. \looseness=-1

To adapt BLiMP to our manipulated languages, we apply the same text transformations to the test sets. We implement a rigorous filtering protocol:
(1) We remove categories where the critical word in the minimal pair is a function word (e.g., Determiner-Noun agreement), as these distinctions may be erased in conditions like \NoFunc.
(2) We remove pairs that become identical after manipulation.
(3) We apply intersection filtering: if a minimal pair is removed in any language condition, it is removed from the evaluation set for \textit{all} models. This ensures that all models are evaluated on the exact same set of underlying syntactic phenomena. A list of excluded categories is in Appendix~\ref{minimalpair}.    \looseness=-1

\subsection{Results.}

\begin{table*}[t]
\small
\centering
\begin{adjustbox}{max width=\linewidth}
\begin{tabular}{l|l|llllllllllllll}
\toprule
Condition & \rotatebox{45}{\parbox{1.5cm}{\textbf{Overall\\$5$-gram}}} & \rotatebox{45}{\parbox{1.5cm}{\textbf{Overall\\Transformer}}} & \rotatebox{45}{S-V Agr} & \rotatebox{45}{Irregular} & \rotatebox{45}{NPI} & \rotatebox{45}{Island} & \rotatebox{45}{Filler.Gap} & \rotatebox{45}{Ellipsis} & \rotatebox{45}{Ctrl.Rais} & \rotatebox{45}{Binding} & \rotatebox{45}{Arg.Str} & \rotatebox{45}{Ana.Agr} \\
\midrule
\NaturalFunc\  & 55.5 & 72.7 & 68.5 & 89.7 & 55.9 & 56.0 & 65.9 & 79.3 & 72.5 & 73.9 & 74.2 & 94.8 \\
\NoFunc\  & 54.1 \negdelta{-1.4} & 60.7 \negdelta{-12.0} & 54.0 \negdelta{-14.5} & 59.1 \negdelta{-30.6} & 50.3 \negdelta{-5.6} & 50.5 \negdelta{-5.4} & 44.3 \negdelta{-21.6} & 69.9 \negdelta{-9.4} & 63.7 \negdelta{-8.8} & 64.2 \negdelta{-9.8} & 59.4 \negdelta{-14.9} & 94.8 \posdelta{+0.0} \\
\FiveFunc\  & 55.4 \negdelta{-0.1} & 70.9 \negdelta{-1.8} & 58.9 \negdelta{-9.6} & 89.7 \negdelta{-0.1} & 55.0 \negdelta{-0.9} & 54.9 \negdelta{-1.0} & 72.5 \posdelta{+6.6} & 77.1 \negdelta{-2.2} & 68.1 \negdelta{-4.4} & 72.7 \negdelta{-1.2} & 68.8 \negdelta{-5.4} & 95.3 \posdelta{+0.5} \\
\MoreFunc\  & 52.8 \negdelta{-2.7}& 69.7 \negdelta{-3.0} & 65.6 \negdelta{-2.9} & 88.7 \negdelta{-1.0} & 53.3 \negdelta{-2.6} & 56.5 \posdelta{+0.6} & 55.5 \negdelta{-10.4} & 72.6 \negdelta{-6.7} & 70.5 \negdelta{-2.0} & 73.2 \negdelta{-0.8} & 70.9 \negdelta{-3.4} & 94.0 \negdelta{-0.8} \\
\BigramDep  & 56.1 \posdelta{+0.6} & 67.4 \negdelta{-5.3} & 56.2 \negdelta{-12.3} & 85.9 \negdelta{-3.8} & 54.6 \negdelta{-1.2} & 50.6 \negdelta{-5.4} & 67.2 \posdelta{+1.3} & 70.2 \negdelta{-9.1} & 66.6 \negdelta{-5.9} & 66.7 \negdelta{-7.3} & 66.1 \negdelta{-8.1} & 93.4 \negdelta{-1.4} \\
\RandDep  & 53.4 \negdelta{-2.1} & 67.0 \negdelta{-5.7} & 56.0 \negdelta{-12.5} & 84.6 \negdelta{-5.2} & 53.1 \negdelta{-2.8} & 52.6 \negdelta{-3.3} & 60.8 \negdelta{-5.1} & 74.9 \negdelta{-4.5} & 65.3 \negdelta{-7.2} & 67.7 \negdelta{-6.3} & 64.1 \negdelta{-10.2} & 95.4 \posdelta{+0.6} \\
\CrossBound  & 54.5 \negdelta{-1.0}& 69.7 \negdelta{-3.0} & 67.1 \negdelta{-1.4} & 83.8 \negdelta{-6.0} & 47.2 \negdelta{-8.6} & 51.5 \negdelta{-4.5} & 67.8 \posdelta{+1.8} & 72.8 \negdelta{-6.6} & 70.7 \negdelta{-1.8} & 70.1 \negdelta{-3.8} & 73.4 \negdelta{-0.9} & 95.8 \posdelta{+1.1} \\

\bottomrule
\end{tabular}
\end{adjustbox}
\caption{BLiMP accuracy by training condition, averaged over 3 seeds. Deltas are relative to \NaturalFunc.}
\label{tab:blimpcat}
\end{table*}


Results are shown in Table~\ref{tab:blimpcat}. Note, we focus on which distributional properties support better syntactic generalization, rather than on whether models acquire full human-like syntax. First, $n$-gram models consistently lag behind transformers across all conditions, and improve on \BigramDep~ over \NaturalFunc. The increased scores for Transformers suggest that they capture aspects of syntactic knowledge that are not available to linear heuristic models.\footnote{We additionally evaluate several interaction conditions for both transformer and n-gram models (i.e., \FiveFunc\_\RandDep, \FiveFunc\_\CrossBound, \BigramDep\_\CrossBound). The two model classes show different sensitivity to these combined manipulations, further suggesting that transformers generalize beyond linear heuristic rules. Details are explained in Appendix~\ref{interaction}.} Among the transformer models, \NaturalFunc~ achieves the highest accuracy, whereas \NoFunc~ performs worst, with the other conditions falling in between. A linear mixed-effects model (\texttt{acc $\sim$ condition + (1 $\vert$ category:phenomenon) + (1 $\vert$ seed)}) shows that, relative to \NaturalFunc, all conditions have significant negative effects ($p < 0.05$), except for \FiveFunc, which is marginally worse than \NaturalFunc~($p = 0.08$). These results suggest that disrupting any of the statistical properties of function words makes syntactic generalization harder to learn.

Additionally, we observe a Goldilocks effect in function word distributions: optimal learnability requires function words to be frequent enough to be reliable, yet sufficiently diverse to remain structurally informative. Collapsing the function word inventory into a small set of highly frequent items, as in \FiveFunc, reduces the distinctiveness of these cues and leads to degraded learning despite preserving high frequency and structural association. Conversely, expanding the inventory excessively---which diversifies the inventory but reduces type frequency of function words, as in \MoreFunc---also harms learnability. These results suggest that successful grammatical learning requires a balance between the frequency and diversity of function words.

Effects for individual BLiMP categories are mostly negative, but not uniform. The large improvement for Filler Gap tests, especially in our \FiveFunc~setting, can be attributed to the fact that some of our manipulations simplify the dependencies between the filler and gap by inventories for fillers, which are function words. 

\paragraph{Summary.}
We show that languages in which function words have all three properties are most learnable by LMs and not all distributional properties of function words contribute equally to learnability. Specifically, disrupting the structural association leads to a larger performance drop than manipulating lexical frequency or positional boundary. We also observe a Goldilocks effect in function word distribution: optimal learnability arises when function words are sufficiently frequent to be reliable, yet sufficiently diverse to support distinctions among grammatical structures.

\section{How Are Function Words Represented and Deployed?}
\begin{figure*}[th]
    \centering
    \includegraphics[width=\linewidth]{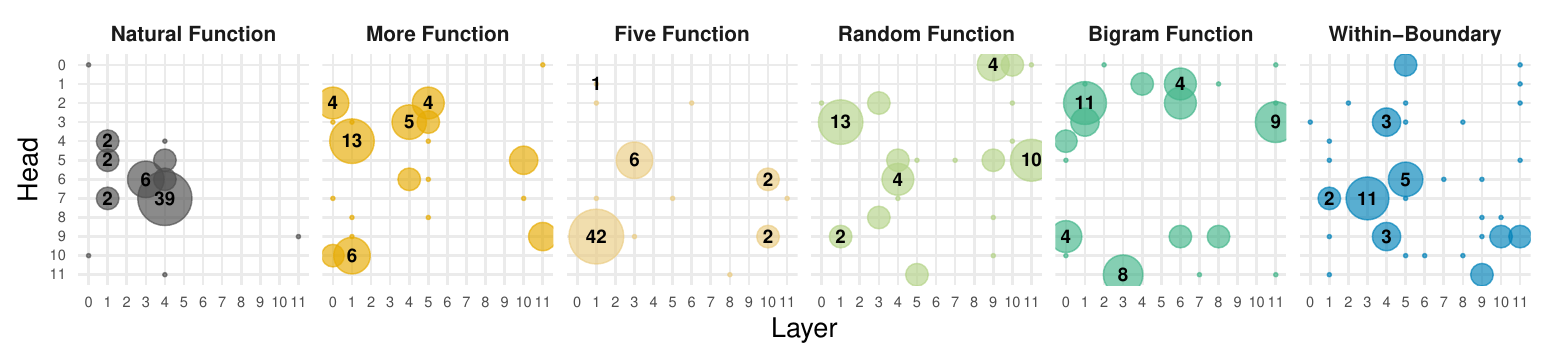}
\caption{Distribution of dominant function heads across BLiMP categories under different training conditions (seed=53). Numbers indicate the top-5 heads and the number of BLiMP categories in which each head assigns the highest attention to function words. Results for other seeds are reported in Appendix~\ref{fig:attresults-other-seeds}.}
    \label{fig:attresults}
\end{figure*}

\begin{figure*}[t]
    \centering
    \begin{subfigure}[t]{0.48\linewidth}
        \centering
        \includegraphics[width=\linewidth]{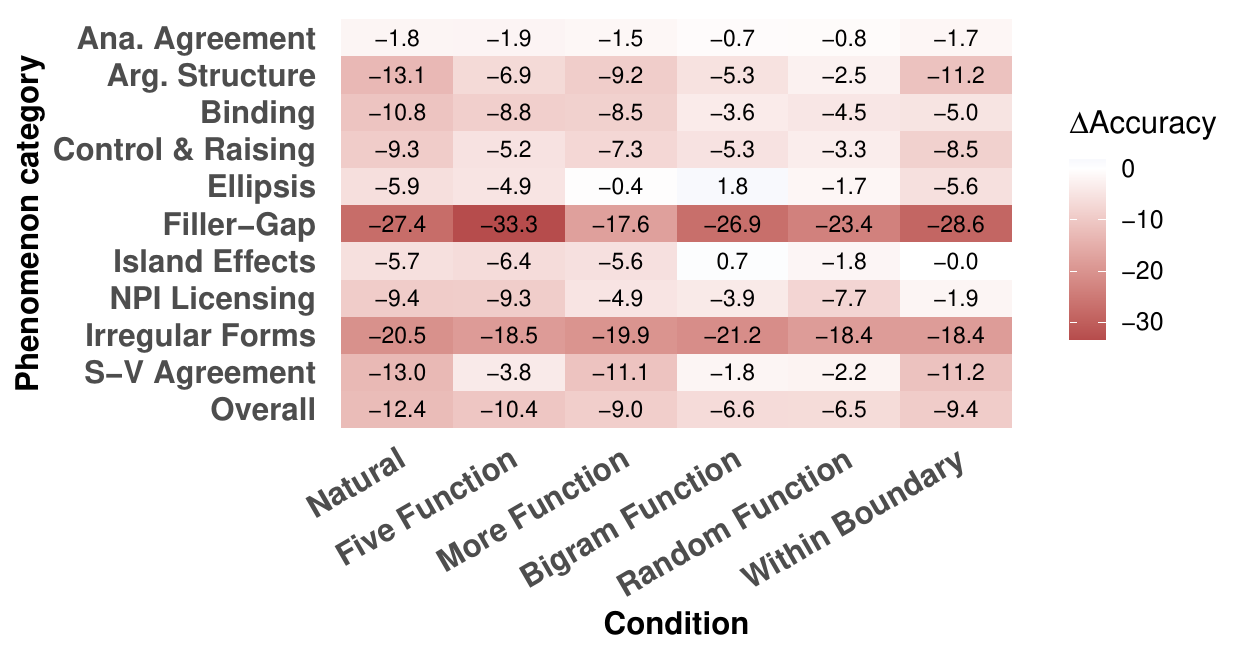}
        \caption{Function word deletion}
        \label{fig:blimptrain}
    \end{subfigure}
    \hfill
    \begin{subfigure}[t]{0.48\linewidth}
        \centering
        \includegraphics[width=\linewidth]{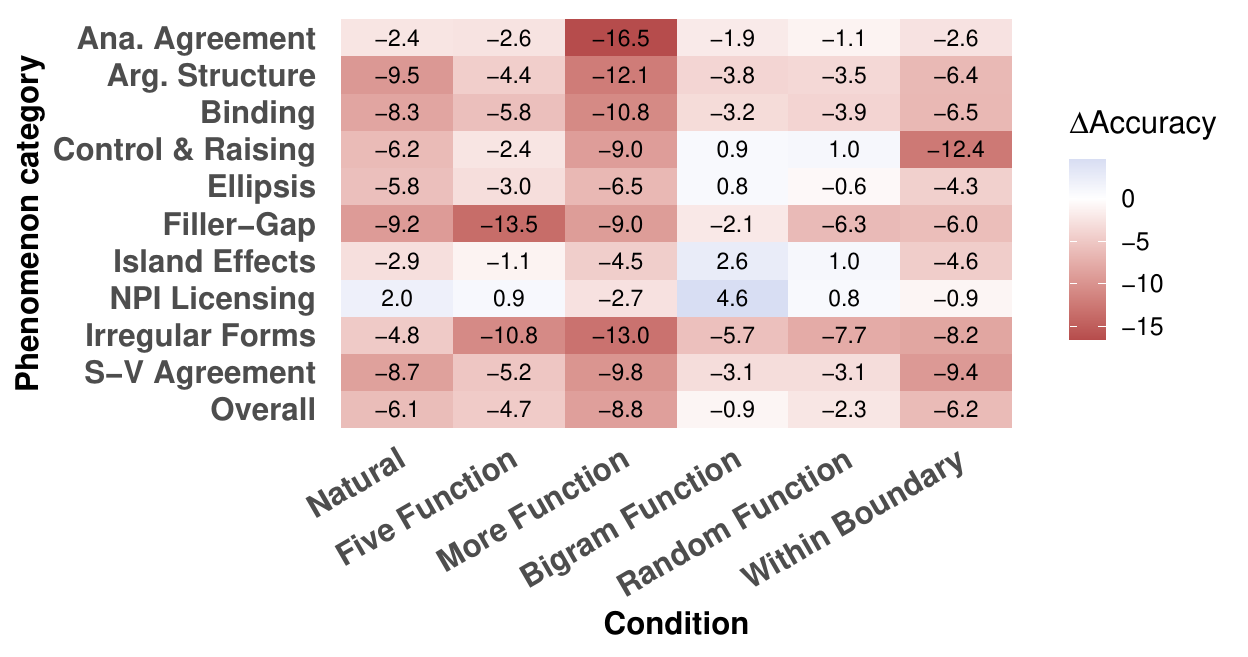}
        \caption{Function word masking}
        \label{fig:blimpinfer}
    \end{subfigure}

    \caption{Difference in BLiMP accuracy before and after ablation.}
    \label{fig:functionwordablation}
    
\end{figure*}
The previous experiments establish that function word distributions shape grammatical learning in our neural models. But do models actually rely on function word information when making grammaticality judgments? If we assume that information actively used at processing time reflects what was encoded during training, then demonstrating processing-time dependence on function words would provide convergent evidence that these distributional properties genuinely drove acquisition rather than serving as incidental correlates of other statistical patterns in the data. We therefore combine attention probing and function word ablation to test this directly.

\subsection{Probing Function Word Attention}
\label{head}
\paragraph{Experiment Setup.} We probe attention patterns to examine whether successful learning is associated with specialized representations for function words.
We apply the probing method of \citet{aoyama-wilcox-2025-language} to identify attention heads that consistently attend to function words across BLiMP categories (see Appendix~\ref{attentionprobing} for details). Models trained without function words (\NoFunc) are excluded from this analysis.

\paragraph{Results.} Figure~\ref{fig:attresults} shows a clear divergence across conditions. 
Models trained with \NaturalFunc\ exhibit a highly concentrated pattern of function word attention: a small number of heads, primarily in Layers 3 and 4, account for the majority of function word attention across BLiMP categories.
This pattern indicates the emergence of specialized representations tied to function words when all three distributional properties are present. In contrast, conditions that disrupt structural predictability (\RandDep~or \BigramDep) or phrase boundaries (\CrossBound) do not produce similarly concentrated representations. The \FiveFunc\ condition shows a more concentrated pattern, likely because of the limited function vocabulary (n=5). Attention to function words in these models is diffuse, suggesting a link between how these words are represented internally and the grammatical abilities of the models.

To confirm that this observation is not due to chance, we visualize the function heads from two additional random seeds in Appendix~\ref{functionhead}. We further aggregate the number of function heads across all heads and layers, and compute the entropy of this distribution across three random seeds. The results are reported in Table~\ref{entropy} in Appendix~\ref{per_seed}. We find that \NaturalFunc~ exhibits lower mean entropy and standard deviation, indicating that function word representations are concentrated in a small number of heads and that this pattern is stable across random seeds.

\subsection{Ablating function word Information}
\paragraph{Experiment Setup.} If models have learned to exploit function word information, then disrupting this information at evaluation time should selectively impair performance, depending on how strongly models rely on such cues. We therefore experiment with two types of ablations which we call function word \textit{masking} and function word \textit{deletion}. In function word masking, we block attention \textit{to} and \textit{from} function word tokens in their corresponding BLiMP benchmark during evaluation, thus turning them into content-free placeholders. In function word deletion, we remove function words entirely from the input by evaluating on \NoFunc\ BLiMP. Models trained without function words (\NoFunc) are excluded from this analysis.

\paragraph{Results.}
Results are summarized in Figure~\ref{fig:functionwordablation}. Overall, function word deletion leads to a larger drop than function word masking, meaning that positional information of function words are helpful for LMs to process texts. In the function word deletion experiment, models trained on \NaturalFunc~show the largest performance drop, followed by \FiveFunc~and \CrossBound, indicating that these models rely heavily on function word information for making judgments. The \MoreFunc~condition leads to the largest drop in the function word masking experiment, likely because the increased number of function words creates stronger associations between more diverse function words and specific structural contexts, making their identity particularly informative. In contrast, \BigramDep~and \RandDep~show the smallest performance drops for both experiments. This suggests that the association between function words and their underlying structural roles is disrupted, models become less reliant on function words.\looseness=-1

For both experiments, we perform paired $t$-tests comparing each ablated model with its corresponding non-ablated baseline. In the deletion experiment, all conditions show significant drops ($p < 0.01$). In the masking experiment, performance decreases significantly in most conditions ($p < 0.05$), except for \BigramDep~and \MoreFunc. However, the masking results are less stable across runs, indicating that models do not consistently rely on function word identity under this manipulation.

\paragraph{Summary.}  Models trained on different language variants differ systematically in how they represent and deploy function words. Disrupting any of the three properties prevents the emergence of focused function word attention heads and decrease the reliance on function words. In particular, disrupting reliable structural association leads to weaker reliance on function words overall. 

\section{Discussion}
\label{7}
Beyond replicating earlier findings from natural language acquisition
\citep[e.g.,][]{valian1988anchor,morgan1987structural},
our computational approach extends prior artificial-language-based studies to more complex naturalistic text and refines existing function word related hypotheses. We show that not all statistical cues contribute equally to grammatical learning in neural learners. In particular, high frequency and reliable structural association exert stronger effects on both learning outcomes and internal representations than
phrase-boundary alignment. 

A natural follow-up question is why these properties matter and why they exhibit a graded pattern of influence. Here we offer a tentative interpretation, grounded in the prior literature on linguistic typology and language acquisition.

We first propose that these properties matter because they maximize learnability. Cognitive scientists have argued that aspects of language structure may result from the particular ways in which humans learn languages \citep{christiansen2008language, culbertson2012learning, newport2016statistical}. According to these proposals, languages are shaped by selectional pressures from successive generations of learners, with learnability being one key selective pressure among others (e.g., communication, compression, see \citealp{kirby2015compression,kirby1999function}). Languages that are easier to learn are more likely to be transmitted successfully across generations. 

With the same amount of input, language learnability emerges from the interaction between learner-internal factors and language-specific properties. Languages can evolve structural features that facilitate learning, while learners bring inductive biases that privilege such patterns, accelerating acquisition. In our experiment, we treat transformer LMs as weakly biased, domain-general statistical learners and use them to re-examine claims from the acquisition literature under controlled conditions. Our results show that languages are most learnable by neural learners when they exhibit all three function word properties. Critically, all three properties are attested in the 186 languages we examined, suggesting they represent convergent evolutionary solutions to the learnability problem. This claim about learnability aligns with studies of human learners, who tend to focus distributional analysis on function words and learn the predictable syntactic distribution of a function word more easily than the equally predictable distribution of a content word \cite{getz2019privileged}. This learning asymmetry aligns with the statistical asymmetry we observe: function words reliably predict syntactic structure, making them effective anchors for grammar learning. Therefore, both the statistical regularities in natural language and human inductive biases contribute to learning efficiency. \looseness=-1

Second, we address why these three properties facilitate learning in the first place, and why they exhibit the observed graded pattern of influence. Following \citet{gerken1987telegraphic,gerken1993interplay}, a long-standing proposal in the acquisition literature distinguishes two foundational challenges prior to syntactic analysis: segmentation, identifying constituent boundaries in the input, and labeling, determining the syntactic type of those constituent types.\footnote{Related distinctions have also been noted in earlier theoretical work \citep{fodor1967some,ClarkClark1977}.} Viewed through this lens, the three distributional properties we study may support learning in distinct ways. High frequency makes function words salient and trackable, facilitating their availability as learning cues, which may help explain why Zipfian distributions support language learning and generalization \citep[e.g.,][]{lavi2022learnability,wolters2024zipfian}. Phrase-boundary alignment plausibly contributes to segmentation by providing positional cues in linear input. Reliable structural association, by contrast, supports labeling by consistently linking function words to particular constituent types, consistent with evidence that function words cue grammatical category learning \citep{mintz2003frequent,zhang2015grammatical}. Consistent with this interpretation, we find that disrupting structural association leads to larger learning costs than disrupting boundary alignment, suggesting that labeling-related information plays a more central role for function words, while segmentation cues may be partially recovered from other sources, such as transition statistics. In summary, by providing the segmentation and labeling information, function words thus serve as anchors to specific syntactic structures and categories, bootstrapping grammar learning.  

\section{Conclusion}

In this study, we investigated how function words contribute to syntactic learning using corpus analysis and computational simulations. Building on generalizations about function words originally discovered through linguistic analysis, we identified three distributional properties of function words (i.e., high frequency, reliable structural association, and systematic alignment with phrase boundaries) and showed that they are robustly attested across 186 languages. Using controlled manipulations and probing analyses, we demonstrated that these properties differentially support grammatical learning in transformer LMs. Languages that preserve all three properties are learned most effectively, mirroring classic findings from studies on human language learning. In addition, disrupting individual properties incurs learning costs. Among the properties we studied, reliable structural association shows stronger influences than boundary alignment and frequency, suggesting that not all statistical cues contribute equally to learning. Finally, we interpret these findings through the framework of cultural evolution: function words act as \textit{anchors} that have been selected for during cultural transmission to maximize learnability, allowing learners to bootstrap complex syntactic structure from limited input via the segmentation and labeling information they provide.

\section*{Limitations}
\label{limitation}
First, our modeling experiments focus exclusively on English. 
While we conduct a cross-linguistic analysis across 186 languages to argue for the universality of function word properties, the counterfactual modeling experiments are limited to English 
for methodological reasons. Our experimental design requires: (i) high-quality dependency 
parsing to identify phrase boundaries and implement the \CrossBound\ manipulation, and (ii) a standardized benchmark for evaluating syntactic generalization. English uniquely satisfies 
both requirements with reliable parsing tools (Stanza) and the BLiMP benchmark, enabling the controlled comparisons central to our research questions. Although our cross-linguistic analysis demonstrates that the three distributional properties hold robustly across typologically diverse languages, testing whether the absolute magnitude of these effects varies with language-specific factors (e.g., word order, morphological richness) remains an important direction for future work.

Second, all of our experiments focus on word-level function categories and do not consider grammatical morphology. In many languages, like Turkish, function-like information is encoded in bound morphemes rather than free function words. UD also ignores the morphology level annotation. Investigating whether and how such morphological markers support syntactic learning and inference would be an important extension of the present study. 

Third, prior work in language acquisition emphasizes the role of prosodic cues, such as stress and rhythm, in helping learners distinguish function words from content words \citep[e.g.,][]{morgan1987structural}. Our experiments, however, rely exclusively on written text. It remains to be seen whether the same patterns hold when models are trained on speech input, where prosodic information may provide additional structural cues.

Fourth, the language \MoreFunc~ introduces potential randomness for each function word, as each pseudo-function word corresponding to a function word is randomly assigned. This design disrupts the natural predictive associations between function words and content words found in actual language. Because these pseudo-function words lack natural collocational histories, it is unclear how they would associate with particular syntactic structures in a realistic linguistic setting. However, this does not mean that \MoreFunc is similar to \RandDep. In \MoreFunc, the pseudo words are based on the original function word, which means that if a pseudo word appears in one context associated with one function word, it cannot appear in another context. We believe that despite this simplification, the current formulation still provides useful insights and serves as a first step toward understanding the role of function words in structured representations.

Finally, it is inherently difficult to determine whether our experimental setup, i.e., evaluating LMs' syntactic generalization using minimal pairs, captures genuine syntactic knowledge like humans or reliance on other heuristic strategies. While BLiMP covers a broad range of linguistic phenomena and the test sentences are distinct from the training data, such evaluations cannot fully rule out alternative explanations based on surface-level cues. For example, a $5$-gram model can reach above-chance accuracy as shown in our experiment. Therefore, we treat BLiMP as a useful \textbf{proxy} for syntactic generalization in LMs, while acknowledging that success on this benchmark does not imply that models acquire grammatical knowledge in the same way as humans.


\section*{Ethical Considerations}
Wikipedia data are publicly available and released under open licenses, including the GNU Free Documentation License (GFDL) and the Creative Commons Attribution--ShareAlike 4.0 License (CC BY-SA 4.0). The UD treebanks used in this study are released under the CC BY-SA 4.0 license. None of the data contain personally identifiable information. For certain UD treebanks, we exclude them due to copyright restrictions. We do not anticipate any potential risks associated with our experiments.

\section*{Acknowledgements}
We thank Elissa Newport, Najoung Kim, and members of NERT lab for their helpful discussion and suggestions. Special thanks to anonymous reviewers and AC for their constructive feedback.

\bibliography{custom}
\appendix
\section{BLiMP Categories Considered}
\label{minimalpair}
We list the categories that are removed from our experiments because the function word is the key to making a correct judgment. The list is reported in Table~\ref{tab:phenomenon-subcategories}.

\section{Training Data Statistics and Training Details}

\label{trainingdetails}
\label{stats}
\begin{table}[h]
\small
\centering
\begin{tabular}{l r}
\toprule
\textbf{Statistic} & \textbf{Value} \\
\midrule
Average sentence length & 25.34 \\
Total number of sentences & 3,519,842 \\
Total number of words & 89,192,973 \\
\bottomrule
\end{tabular}
\caption{Overall corpus statistics.}
\label{tab:corpus-overall}
\end{table}

\begin{table}[h]
\small
\centering
\begin{tabular}{l r}
\toprule
\textbf{Split} & \textbf{Number of Words} \\
\midrule
Training set & 76,007,866 \\
Development set & 108,728 \\
Test set & 217,241 \\
\bottomrule
\end{tabular}
\caption{Train, development, and test split sizes in number of words.}
\label{tab:corpus-split}
\end{table}

\begin{table}[h]
    \centering
    \small
\begin{tabular}{lr}
\toprule
\multicolumn{2}{c}{Model}\\
\midrule
\texttt{model\_type} & \texttt{GPT2-small}\\
\texttt{seeds} & \texttt{42,53,67}\\
\midrule
\multicolumn{2}{c}{Tokenizer}\\
\midrule
\texttt{vocab\_size} & \texttt{32,768}\\
\texttt{tokenization\_alg} & BPE\\
\texttt{training\_data} & Wiki train split\\
\midrule
\multicolumn{2}{c}{Training}\\
\midrule
\texttt{num\_epoch} & \texttt{10}\\
\texttt{batch\_size} & \texttt{128}\\
\texttt{context\_length} & \texttt{128}\\
\texttt{grad\_acc\_steps} & \texttt{1}\\
\texttt{weight\_decay} & \texttt{0.1}\\
\texttt{warmup\_steps} & \texttt{10\%}\\
\texttt{lr} & \texttt{5e-4}\\
\texttt{lr\_scheduler} & \texttt{linear}\\
\texttt{GPU} & \texttt{Tesla V100}\\
\bottomrule
\end{tabular}
\caption{Model and training parameters}
\end{table}

\section{Function Word Inventory}
\label{funcw}
Function words are collected from the gold annotations of GUM and EWT with manual corrections. We also filter function word pieces that are less frequent than 10 times (e.g., \textit{wilt}).  
\begin{table}[h]
\centering
\small
\setlength{\tabcolsep}{6pt}
\begin{tabular}{lp{0.70\linewidth}}
\toprule
\textbf{Category} & \textbf{Function Words} \\
\midrule
DET & the, this, a, an, no, all, another, each, that, any, those, these, both, every, either, neither \\[4pt]

CCONJ & and, but, or, yet \\[4pt]

SCONJ & that, if, although, after, whereas, while, before, as, though, until, because, since, once, whether, unless, albeit, till, whilst \\[4pt]

AUX & will, be, had, were, being, is, would, was, do, could, are, have, been, has, did, should, might, can, does, ’s, may, must, ca, am, shall, art, ar, re, ought, need \\[4pt]

ADP & at, in, of, near, for, by, to, with, on, from, behind, into, within, despite, against, as, over, than, during, about, between, among, except, through, around, after, like, off, without, under, before, throughout, unlike, across, toward, along, above, aboard, until, upon, via, beneath, unto, beyond, per, below, amongst, till, beside, amid, onto, towards, underneath, alongside \\
\bottomrule
\end{tabular}
\caption{Function word inventories used in the experiments, grouped by syntactic category.}
\label{tab:functionwords}
\end{table}
\section{Manipulated Properties in Each Language}
\label{condition}
See Table~\ref{tab:conditions}.

\begin{figure*}[th]
    \centering

    \begin{subfigure}{\linewidth}
        \centering
        \includegraphics[width=\linewidth]{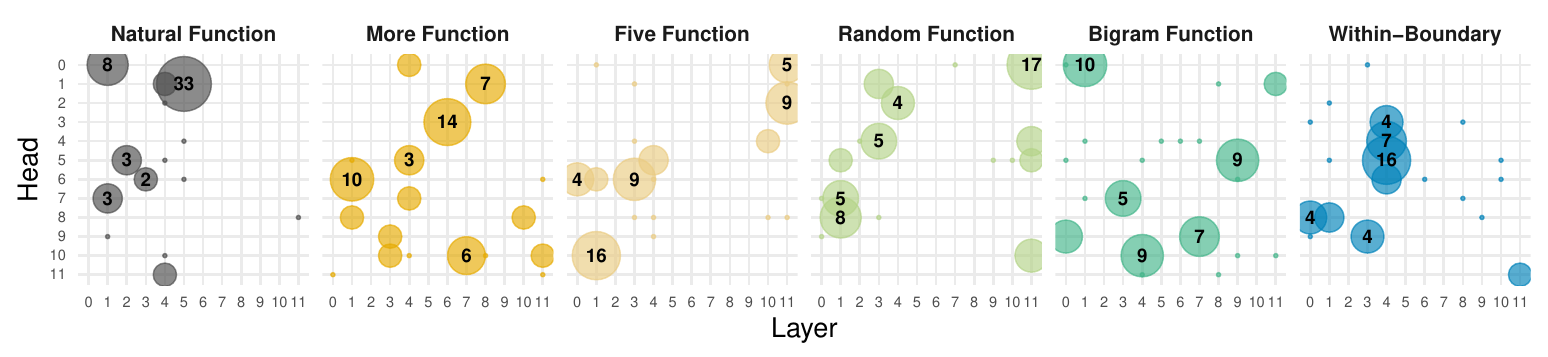}
        \caption{Random seed = 42}
        \label{fig:function-head-42}
    \end{subfigure}

    \vspace{1em}

    \begin{subfigure}{\linewidth}
        \centering
        \includegraphics[width=\linewidth]{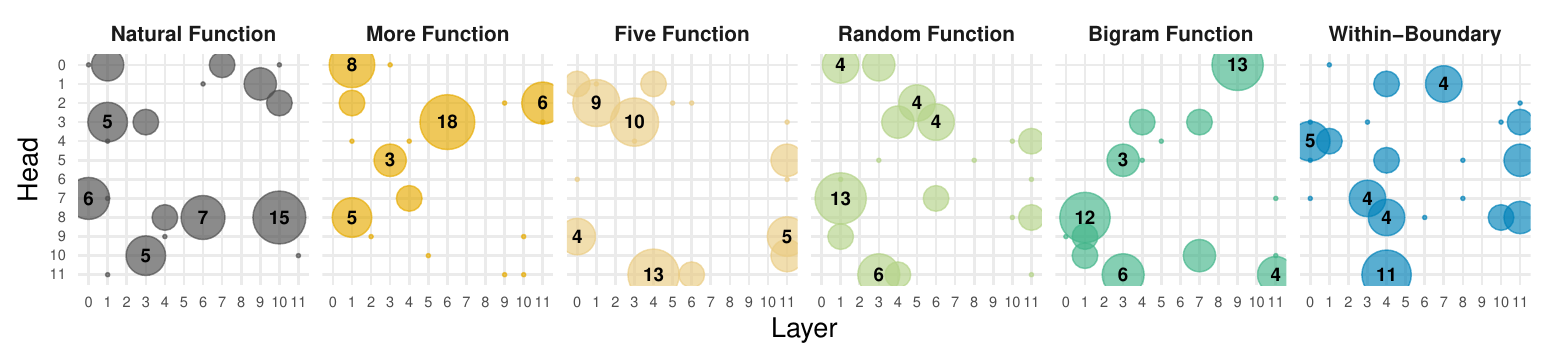}
        \caption{Random seed = 67}
        \label{fig:function-head-67}
    \end{subfigure}

    \caption{Distribution of the dominant function head across BLiMP categories under different training conditions.
    Numbers in each subplot indicate the top-5 most frequent heads and the number of categories for which each head assigns the highest attention to function words.}
    \label{fig:attresults-other-seeds}
\end{figure*}

\begin{table*}[th]
\small
    \centering
    \begin{tabular}{l|ccccccccc}
    \toprule
    Language  & Lexical Frequency & Structural Association & Phrase Boundary\\
    \midrule 
    \NaturalFunc &  \textcolor{green}{\Checkmark} high & \textcolor{green}{\Checkmark} preserved & \textcolor{green}{\Checkmark} aligned \\
    \NoFunc   & \textcolor{red}{\XSolidBrush}  zero & \textcolor{red}{\XSolidBrush}  none & \textcolor{red}{\XSolidBrush}  none \\
    \FiveFunc &  \textcolor{green}{\Checkmark} \textcolor{green}{\Checkmark} very high & \textcolor{green}{\Checkmark} preserved& \textcolor{green}{\Checkmark} aligned\\
    \MoreFunc & \textcolor{red}{\XSolidBrush}  low & \textcolor{green}{\Checkmark} preserved & \textcolor{green}{\Checkmark} aligned\\
    \BigramDep & \textcolor{green}{\Checkmark} high& \textcolor{red}{\XSolidBrush}  destroyed & \textcolor{green}{\Checkmark} aligned \\
    \RandDep &  \textcolor{green}{\Checkmark} high & \textcolor{red}{\XSolidBrush} destroyed & \textcolor{green}{\Checkmark} aligned\\
    \CrossBound  & \textcolor{green}{\Checkmark} high & \textcolor{green}{\Checkmark} preserved & \textcolor{red}{\XSolidBrush}  disrupted\\
    \midrule
    \midrule 
    \FiveFunc\_\CrossBound & \textcolor{green}{\Checkmark} \textcolor{green}{\Checkmark} very high & \textcolor{green}{\Checkmark} preserved & \textcolor{red}{\XSolidBrush} disrupted \\

    \FiveFunc\_\RandDep & \textcolor{green}{\Checkmark} \textcolor{green}{\Checkmark} very high & \textcolor{red}{\XSolidBrush} destroyed & \textcolor{green}{\Checkmark} aligned\\

    \BigramDep\_\CrossBound &   \textcolor{green}{\Checkmark} high & \textcolor{red}{\XSolidBrush} destroyed & \textcolor{red}{\XSolidBrush} disrupted\\
    
    \bottomrule
    \end{tabular}
    \caption{The manipulated properties in different counterfactual languages.}
    \label{tab:conditions}
\end{table*}

\begin{table}[t]
\centering
\small
\begin{adjustbox}{max width=\linewidth}
\begin{tabular}{ll}
\toprule
\textbf{Phenomenon group} & \textbf{Sub-phenomenon} \\
\midrule
\multirow{8}{*}{\sout{Determiner--noun agreement}}
 & \sout{determiner\_noun\_agreement\_1} \\
 & \sout{determiner\_noun\_agreement\_2} \\
 & \sout{determiner\_noun\_agreement\_irregular\_1} \\
 & \sout{determiner\_noun\_agreement\_irregular\_2} \\
 & \sout{determiner\_noun\_agreement\_with\_adjective\_1} \\
 & \sout{determiner\_noun\_agreement\_with\_adjective\_2} \\
 & \sout{determiner\_noun\_agreement\_with\_adj\_irregular\_1} \\
 & \sout{determiner\_noun\_agreement\_with\_adj\_irregular\_2} \\

\midrule
\multirow{1}{*}{NPI licensing}
 & \sout{matrix\_question\_npi\_licensor\_present} \\
\midrule
\multirow{4}{*}{Quantifiers}
 & \sout{existential\_there\_quantifiers\_1} \\
 & \sout{existential\_there\_quantifiers\_2} \\
 & \sout{superlative\_quantifiers\_1} \\
 & \sout{superlative\_quantifiers\_2} \\

\bottomrule
\end{tabular}
\end{adjustbox}
\caption{BLiMP phenomenon groups and their corresponding sub-phenomena removed in our experiments (13 subcategories in total).}
\label{tab:phenomenon-subcategories}
\end{table}
\section{Attention Probing Details}
\label{attentionprobing}

Following \citet{aoyama-wilcox-2025-language}, we adapt attention probing techniques originally proposed by \citet{clark-etal-2019-bert} and \citet{chen2024sudden} to autoregressive models such as GPT-2. In contrast to prior work, we focus specifically on dependency relations involving function words.

For a given attention head $h$ at layer $l$, we define a head-specific probe $f_{h,l}$ that predicts the syntactic parent of a target word $x_i$ by selecting the word $x_j$ receiving the strongest attention connection with $x_i$, excluding self-attention:
\begin{equation}
f_{h,l}(x_i) = \arg\max_{j \neq i} a_{ij}^{(h,l)} ,
\end{equation}
where $a_{ij}^{(h,l)}$ denotes the word-level attention strength between $x_i$ and $x_j$, defined as the maximum of attention from $x_i$ to $x_j$ and from $x_j$ to $x_i$ for head $h$ at layer $l$.

Because we use BPE tokenization, we convert token-level attention weights to word-level attention scores. When attending \emph{to} a split word, we sum attention weights over its constituent tokens; when attending \emph{from} a split word, we average attention weights across its tokens.

We quantify the extent to which a given head attends to function words using a \emph{function attention score}. Let $F$ denote the set of function words and let $N$ be the number of evaluated target words. The function attention score for head $h$ at layer $l$ is defined as:
\begin{equation}
\text{S}_F(h,l) =
\frac{1}{N}
\sum_{i=1}^{N}
\mathbb{I}\!\left[f_{h,l}(x_i) \in F
\right] .
\end{equation}

Finally, we identify the specific head $(h^*, l^*)$ that maximizes this function attention score for each subcategory in BLiMP.

\section{Function Head Distribution Results for the Other Two Random Seeds}
\label{functionhead}
See Figure~\ref{fig:attresults-other-seeds}.

\section{Language learning results}
\label{language_competence}
The result is reported in Table~\ref{language_competence}. These results demonstrate that for the majority of conditions, the LMs are highly competent at their respective languages, indicating that their performance drops on BLiMP to a failure in syntactic induction rather than a lack of general language modeling competence. For \RandDep~ and \MoreFunc, the low scores indicate these variants are significantly harder to learn as a language. 

Assigning high probability to in-domain data does not necessarily guarantee good syntactic generalization. What is interesting is that \FiveFunc~ shows the highest score, meaning the language model learns such a distribution the best. However, as shown in Table~\ref{tab:blimpcat}, \FiveFunc~ does not lead to the best BLiMP generalization.

\begin{table}[t]
\small
\centering
\begin{tabular}{lc}
\toprule
\textbf{Language} & \textbf{Score} \\
\midrule
\NaturalFunc   & 0.9681 \\
\NoFunc        & 0.9748 \\
\MoreFunc      & 0.0634 \\
\FiveFunc      & 0.9919 \\
\BigramDep    & 0.9885 \\
\RandDep   & 0.1602 \\
\CrossBound    & 0.9650 \\
\bottomrule
\end{tabular}
\caption{Performance scores across different language conditions.}
\label{tab:lang_performance}
\end{table}
\section{Interaction effect of the statistical properties of function words}
\label{interaction}
\begin{figure}
    \centering
    \includegraphics[width=\linewidth]{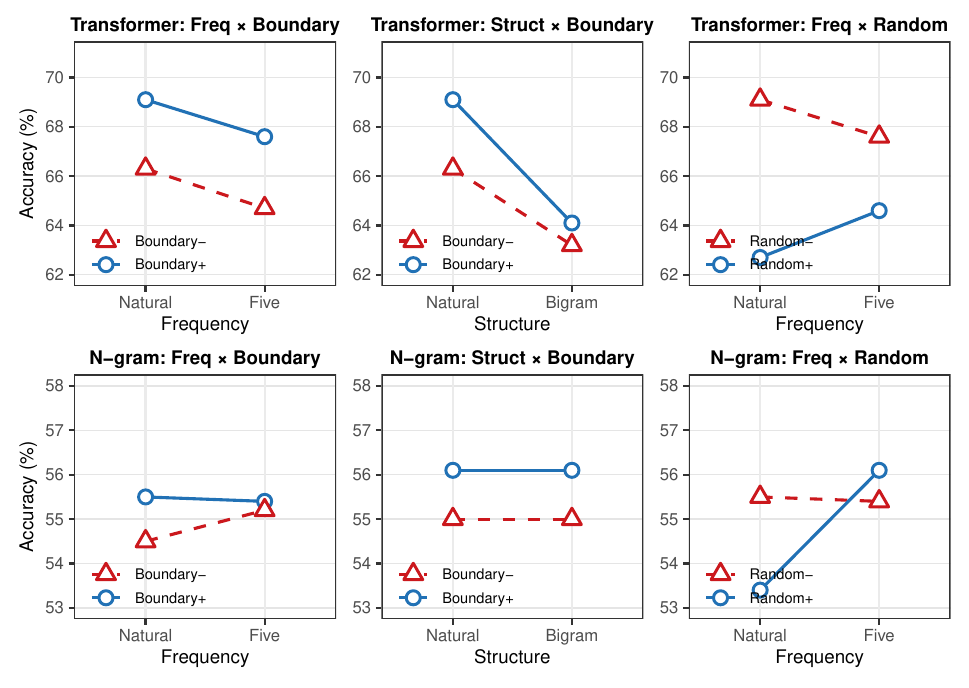}
    \caption{The interaction among the tree properties}
    \label{fig:interaction}
\end{figure}

We further train languages \FiveFunc\_\CrossBound, \FiveFunc\_\RandDep, and \BigramDep\_\CrossBound~ to examine the interaction between frequency, structure, and boundary positions of function words. For the transformer model, we use the same hyperparameters on random seed 53. We also train 5-gram models with the same settings in the main text. The results are shown in Figure~\ref{fig:interaction}.

First, we find that transformer models are sensitive to all three distributional properties and their interactions. Boundary alignment, frequency, and structural dependency each have clear effects on performance. The effects are largely additive in the frequency condition, but we observe a modest interaction between structure and boundary alignment, with the boundary advantage reduced under the \BigramDep~condition. In addition, randomization has a strong negative effect, particularly under natural frequency distributions compared with \FiveFunc~ condition, indicating that disrupting the function word system is more harmful when distributional cues are informative.

In contrast, n-gram models exhibit minimal sensitivity to these manipulations. Performance remains largely unchanged across frequency and structural conditions, and boundary effects are weak and inconsistent. Only the randomization condition shows some variation, but the pattern differs from that observed in transformer models: a randomized language with only five function words is better than a five-function language without randomization. This contrast suggests that transformer models capture structured dependencies among function words, whereas n-gram models rely primarily on local statistical patterns.

\section{Function head results}
\label{per_seed}

For the attention-head masking experiment, we constructed a 12×12 layer–head frequency matrix for each condition, linearized it, and computed the entropy of the resulting distribution. The averaged results are reported in Table~\ref{entropy} and per-seed result in Table~\ref{tab:seed_results}.

\NaturalFunc achieves the lowest mean entropy overall. We also observe that models trained with seed 67 consistently exhibit higher entropy across all conditions, suggesting relatively sparser function-head representations. 

\begin{table}[!th]
\small
\centering
\begin{tabular}{lcc}
\toprule
\textbf{Condition} & \textbf{Mean Entropy (bits)} & \textbf{SD} \\
\midrule
\NaturalFunc    & \textbf{2.74} & 0.861 \\
\MoreFunc       & 3.53 & 0.263 \\
\FiveFunc       & 2.87 & 0.924 \\
\RandDep        & 3.64 & 0.156 \\
\BigramDep      & 3.60 & 0.257 \\
\CrossBound     & 4.02 & 0.463 \\
\bottomrule
\end{tabular}
\caption{Mean entropy and standard deviation across different conditions. \NaturalFunc has the lowest mean entropy, indicating that its function head distribution is more concentrated.}
\label{entropy}
\end{table}

\begin{table}[!th]
\small
\centering
\begin{tabular}{lccc}
\toprule
Condition & Seed 42 & Seed 53 & Seed 67 \\
\midrule
\NaturalFunc   & \textbf{2.47} & 2.05 & 3.71 \\
\MoreFunc      & 3.57 & 3.78 & 3.26 \\
\FiveFunc      & 3.42 & \textbf{1.81} & 3.40 \\
\RandDep       & 3.47 & 3.69 & 3.77 \\
\BigramDep     & 3.71 & 3.78 & \textbf{3.31} \\
\CrossBound    & 3.55 & 4.47 & 4.03 \\
\bottomrule
\end{tabular}
\caption{Results across different random seeds.}
\label{tab:seed_results}
\end{table}

\begin{table*}[!th]
\small
    \centering
    \begin{adjustbox}{max width=\linewidth}
    \begin{tabular}{l|lll}
    
    \toprule
    Language & Sentence (correct) & Sentence (incorrect)\\
    \midrule 
     \NaturalFunc  & brian thought that rebecca was criticizing him .  &  brian thought that rebecca was criticizing himself .\\
     \NoFunc    &  brian thought rebecca criticizing him . & brian thought rebecca criticizing himself .\\

     \FiveFunc & brian thought the rebecca will criticizing him . & brian thought the rebecca will criticizing himself . \\
     \MoreFunc & brian thought blat rebecca wam criticizing him . & brian thought blat rebecca wam criticizing himself . \\
     \BigramDep & brian thought albeit rebecca with criticizing him & brian thought albeit rebecca with criticizing himself .\\
     \RandDep & brian thought since rebecca until criticizing him . & brian thought since rebecca until criticizing himself .\\
     \CrossBound & brian thought rebecca was that criticizing him .& brian thought rebecca was that criticizing himself .\\
     \midrule
     \midrule 
     \FiveFunc\_\CrossBound & brian thought rebecca that that criticizing him . & brian thought rebecca that that criticizing himself .\\
     \FiveFunc\_\RandDep & brian thought that rebecca that criticizing him . & brian thought that rebecca that criticizing himself .\\
     \BigramDep\_\CrossBound &  brian thought rebecca albeit with criticizing him . & brian thought rebecca albeit with criticizing himself .\\
     \bottomrule
    \end{tabular}
    \end{adjustbox}
    \caption{Minimal pair examples in different created BLiMP benchmarks. Example sentence: \textit{Brian thought that Rebecca was criticizing him.} }
    \label{tab:minimalpair}
\end{table*}

\end{document}